%% file: top.tex
\documentclass[10pt,twocolumn,letterpaper]{article}

\usepackage{cvpr}
\usepackage{times}
\usepackage{epsfig}
\usepackage{graphicx}
\usepackage{amsmath}
\DeclareMathOperator*{\argmax}{argmax}
\usepackage{amssymb}
\usepackage{multirow}
\usepackage[caption=false]{subfig}
\usepackage{tabulary}
\usepackage{epigraph}
\usepackage[table]{xcolor}
\usepackage{pifont}

\usepackage[pagebackref=true,breaklinks=true,letterpaper=true,colorlinks,bookmarks=false]{hyperref}

\newcommand{\bd}[1]{\textbf{#1}}
\newcommand{\bt}[1]{\textbf{\textit{#1}}}

\newcommand{\sbd}[1]{\scriptsize{(\textbf{#1})}}
\newcolumntype{x}[1]{>{\centering\arraybackslash}p{#1pt}}
\newlength\savewidth\newcommand\shline{\noalign{\global\savewidth\arrayrulewidth
 \global\arrayrulewidth 1pt}\hline\noalign{\global\arrayrulewidth\savewidth}}

\newcommand{\thicktilde}[1]{\mathbf{\tilde{\text{$#1$}}}}
\definecolor{grey}{rgb}{0.9,0.9,0.9}

\cvprfinalcopy 

\def\cvprPaperID{} 

\ifcvprfinal\fi
\begin{document}

\title{PnPNet: End-to-End Perception and Prediction with Tracking in the Loop}

\author{
  Ming Liang$^{1}$\thanks{Equal contribution.} \quad Bin Yang$^{1,2 *}$\\
  Wenyuan Zeng$^{1,2}$ \quad Yun Chen$^{1}$ \quad Rui Hu$^{1}$ \quad Sergio Casas$^{1,2}$ \quad Raquel Urtasun$^{1,2}$\\
  $^{1}$Uber Advanced Technologies Group \quad $^{2}$University of Toronto\\
  \small\texttt{\{ming.liang, byang10, wenyuan, yun.chen, rui.hu, sergio.casas, urtasun\}@uber.com}
}

\maketitle
\thispagestyle{empty}

\input{abstract}
\input{intro}
\input{related}
\input{model}
\input{experiments}
\input{conclusion}

{\small
\bibliographystyle{ieee_fullname}
\bibliography{egbib}
}

\input{appendix}

\end{document}

%% file: abstract.tex
\begin{abstract}

We tackle the problem of joint perception and motion forecasting in the context of self-driving vehicles. 
Towards this goal we propose PnPNet, an end-to-end model that takes as input sequential sensor data, and outputs at each time step object tracks and their future trajectories. 
The key component is a novel tracking module that generates object tracks online from detections and exploits trajectory level features for motion forecasting.
Specifically, the object tracks get updated at each time step by solving both the data association problem and the trajectory estimation problem.
Importantly, the whole model is end-to-end trainable and benefits from joint optimization of all tasks.
We validate PnPNet on two large-scale driving datasets, and show significant improvements over the state-of-the-art with better occlusion recovery and more accurate future prediction.

\end{abstract}

%% file: intro.tex

\section{Introduction}
We focus on the task of joint perception and prediction (motion forecasting) in the context of self-driving vehicles. 
This is a crucial task as in order to plan a safe maneuver, anticipating the future decisions of surrounding
agents is as important as estimating their current state.

\begin{figure}[t]
\begin{center}
   \includegraphics[width=1.0\linewidth]{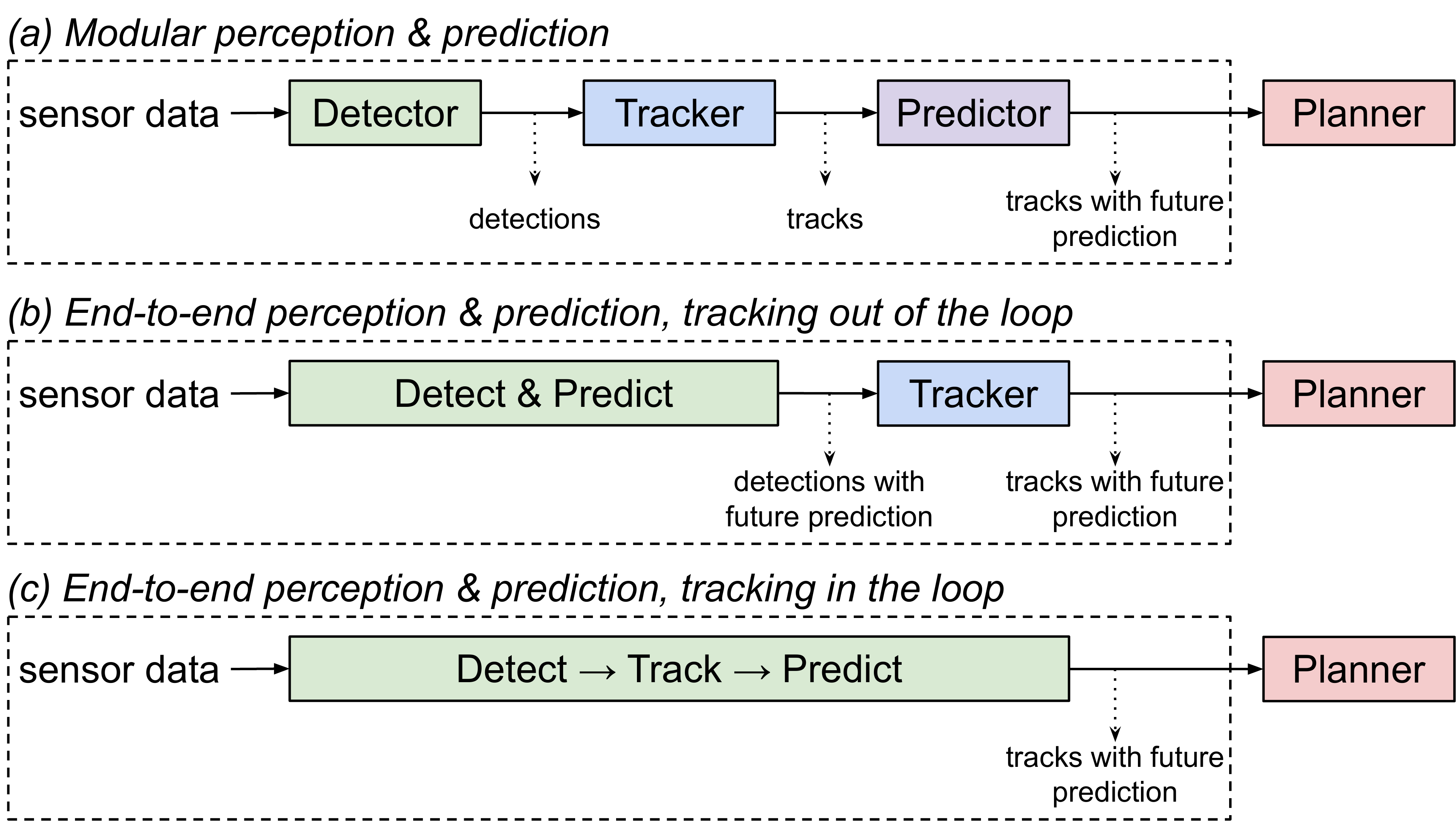}
\end{center}
\caption{\textbf{Three paradigms for perception and prediction.} Traditional approach (a) adopts the modular design that decomposes the stack into subtasks and solves them with individual models. End-to-end method like \cite{dpt} (b) uses a joint model to solve detection and prediction simultaneously, but performs tracking as post-processing. As a result, the full temporal history contained in tracks is not used by detection and prediction. Our approach (c) brings tracking into the loop so that all tasks benefit from rich temporal context.}
\label{fig:intro}
\end{figure}

Different paradigms have been proposed to solve the perception and prediction problem, which are compared in Figure \ref{fig:intro}.
Traditional self-driving autonomy stacks \cite{bansal2018chauffeurnet, chai2019multipath, djuric2018motion} decompose the problem into three subtasks: object detection, object tracking, and motion forecasting, and rely on independent components that perform these subtasks sequentially.
However, as each component is developed separately, this paradigm makes compromises in each module in order to meet the computing budget. 
Furthermore, the interface between these modules is very compact (typically the object's position, velocity, acceleration, and their uncertainty estimates), which prevents downstream tasks from correcting the mistakes made by upstream ones.

Recently, models that solve the detection and prediction tasks jointly with a single neural network have been proposed \cite{dpt}, resulting in more efficient computation and improved accuracy. This paradigm is later extended to further solve the driver intention \cite{intentnet} and motion planning \cite{nmp} by adding the corresponding modules on top of the shared backbone network. These approaches, however, suffer from limited use of temporal history because object tracking is not included in the loop and thus only leverage up to 1 second of past sensor data due to limited model capacity. This may cause problems when dealing with occluded actors and may produce temporal inconsistency in predictions.

In this paper we argue that leveraging the past is key for sequential decision making process like motion forecasting. Towards this goal we propose \textit{PnPNet}, a new paradigm that combines ideas from multi-object tracking and joint perception and prediction models.
While the detection module processes sequential sensor data and generates object detections at each time step, the tracking module associates these estimates across time for better understanding of object states (\eg, occlusion reasoning, trajectory smoothing), which in turn provides richer information for the prediction module to produce accurate future trajectories. Importantly, all modules share computation as there is a single backbone network, and the full model can be trained end-to-end.

We make two main technical contributions in PnPNet. 
First, we propose a novel object trajectory representation defined on a sequence of object detections to fully capture the temporal characteristics of the actors.
In particular, for each object we first extract its inferred motion (from past detection estimates) and raw observations (from sensor features) at each time step, and then model its dynamics using a recurrent network. Importantly, this trajectory representation is utilized in both tracking and prediction modules. 
Second, we propose a  multi-object tracker that solves both the discrete problem of data association and the continuous problem of trajectory estimation \cite{milan2015multi} via  learnable functions that can  handle object occlusion, new birth of trajectories and false positive detections.

We validate PnPNet on two large-scale driving datasets, and demonstrate its effectiveness with both modular metrics (standard benchmark for each subtask) and system metrics (end-to-end performance under the real-world setting).
Experiments show that PnPNet achieves significant improvements over previous state-of-the-art paradigms in both perception and prediction tasks.
Specifically, PnPNet recovers objects from occlusion, produces more complete object trajectories, and outputs more accurate future predictions.

\begin{figure*}[t]
\begin{center}
   \includegraphics[width=1.0\linewidth]{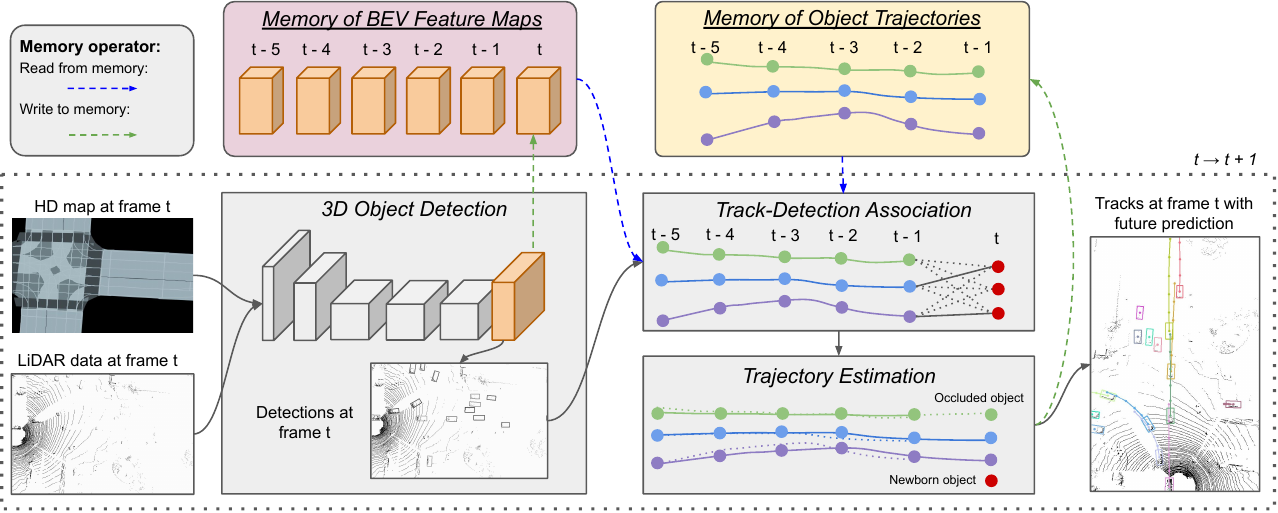}
\end{center}
\caption{\textbf{The proposed PnPNet for end-to-end perception and prediction.} The model consists of three modules that perform 3D object detection, discrete-continuous tracking, and motion forecasting sequentially. To extract trajectory level actor representations used for tracking and prediction, we also equip the model with two explicit memories: one for global sensor feature maps, and one for past object trajectories. Both memories get updated at each time step with up-to-date sensor features and tracking results.}
\label{fig:arch}
\end{figure*}

%% file: related.tex
\section{Related Work}

In this section we review works that tackle the tasks of 3D object detection, tracking, and motion forecasting separately, followed by approaches that tackle these jointly.

\paragraph{3D Object Detection:}
While several approaches \cite{3dop,mono3d,wang2019pseudo} try to perform 3D object detection from images, the inherent depth ambiguity hinders them from being applied in safety-critical applications. Methods that exploit depth sensors (\eg LiDAR) achieve superior performance with various representations of point clouds \cite{pixor,voxelnet,second,shi2019pointrcnn,yang2019std,meyer2019lasernet}. Recently sensor fusion methods \cite{mv3d,pointfuse,contfuse,fpointnet,hdnet,mmf,meyer2019sensor} further push the performance by exploiting complementary information from cameras and/or maps. For efficiency and accuracy, PnPNet utilizes the bird's eye view representation of LiDAR and maps and performs single shot detection. 

\paragraph{Multi-Object Tracking:}
Most approaches mainly follow the tracking-by-detection paradigm \cite{breitenstein2010online}, which comprise the discrete problem of data association and continuous problem of trajectory estimation \cite{milan2015multi}. Many frameworks have been proposed to solve the data association problem: \eg, Markov Decision Processes \cite{xiang2015learning}, min-cost flow \cite{lenz2015followme,frossard2018end,schulter2017deep}, linear assignment problem \cite{sharma2018beyond,weng2019baseline} and graph cut \cite{malcolm2007multi,tang2015subgraph}. 
To handle object occlusion when there's no detection available, hand-crafted heuristics \cite{karunasekera2019multiple} or single-object tracking approach \cite{YanTDE,famnet} has been explored. Apart from the association paradigm, different representations are used to computes the affinity. While \cite{weng2019baseline} exploits the 3D motion clues only, approaches that extract sensor features \cite{zhang2019robust,famnet} typically limit the temporal history to 3 time steps. In contrast, PnPNet solves both discrete and continuous problems, with a long-term trajectory representation that captures both sensor observation and motion clue of actors. 

\paragraph{Motion Forecasting:}
Various approaches have been proposed to model the multi-agent interactions and multi-modal behaviors in motion forecasting. DESIRE \cite{lee2017desire} uses a variational auto-encoder to generate trajectory proposals and refines them based on semantic scene context and interactions between agents. To better model the interactions, game theory is used to formulate the problem \cite{ma2017forecasting}. Social-LSTM \cite{alahi2016social} introduces social pooling to model nearby agents' trajectory patterns, while Social-GAN \cite{gupta2018social} further improves the performance by adding adversarial training. 
In parallel to different predictive models, various input representations are also explored. Besides the past states of actors, sensor features are also explored to provide more context \cite{lee2017desire,liang2019peeking,rhinehart2019precog}.
However, these methods are typically developed on ground-truth object labels, and have generalization issues when applied to noisy detections \cite{rhinehart2019precog}.
In self-driving domain, raster representation in bird's eye view that encodes both the perception output and map information is widely used \cite{bansal2018chauffeurnet,chai2019multipath,djuric2018motion,cui2019multimodal}. In contrast, the prediction module in PnPNet directly reuses the perception features for rich scene context, and also extracts object states explicitly from past object tracks.

\paragraph{Joint Models for Perception and Prediction:}
FAF \cite{dpt} proposes to jointly reason about 3D object detection and motion forecasting by exploiting temporal features from multi-sweep LiDAR point clouds. An efficient bird's eye view representation and network architecture are utilized for real-time inference. IntentNet \cite{intentnet} extends the approach by adding the prediction of high-level intentions of each agent from semantic HD maps. SpAGNN \cite{casas2019spatially} leverages graph neural networks with spatial reasoning to model multi-agent interactions. NeuralMP \cite{nmp} takes one step further by sharing the feature for motion planning with perception and prediction, leading to an end-to-end motion planner.  While all these approaches share the sensor features for detection and prediction, they fail to exploit the rich information of actors along the time dimension. PnPNet addresses this by incorporating online tracking and extracting trajectory-level actor representation to encode long-term history, which in turn improves all tasks.

%% file: model.tex
\section{End-to-End Perception and Prediction}

We introduce \textit{PnPNet} (Figure \ref{fig:arch}), an end-to-end model designed for efficient and accurate joint perception and prediction in the context of autonomous driving.
Instead of designing individual models for each subtask like the traditional engineering stack, we follow the recent advances of joint modeling with shared feature computation \cite{dpt, intentnet}.
However, the main weakness of this paradigm is the limited exploitation of history information.
Since these approaches do not have explicit tracking in the loop, to perform motion forecasting the object's motion history has to be estimated from the raw sensor data, which can be particularly difficult for occluded objects. 
As a result, the performance of the model usually saturates with fewer than 1 second of sensor data \cite{dpt, nmp}.
Furthermore, these approaches cannot track through occlusions longer than the input time horizon, as there's no evidence.
All these drawbacks hinder the performance of these approaches in the task of motion forecasting.

In contrast, PnPNet addresses the issue with two key components: a novel trajectory level representation that captures the rich temporal characteristics of actors, and a new online discrete-continuous tracking module that generates such trajectories from detections across time.
In the remainder we first present the three modules that perform detection, tracking and prediction sequentially, and then show how the full model can be trained end-to-end. 

\begin{figure*}[t]
\begin{center}
   \includegraphics[width=1.0\linewidth]{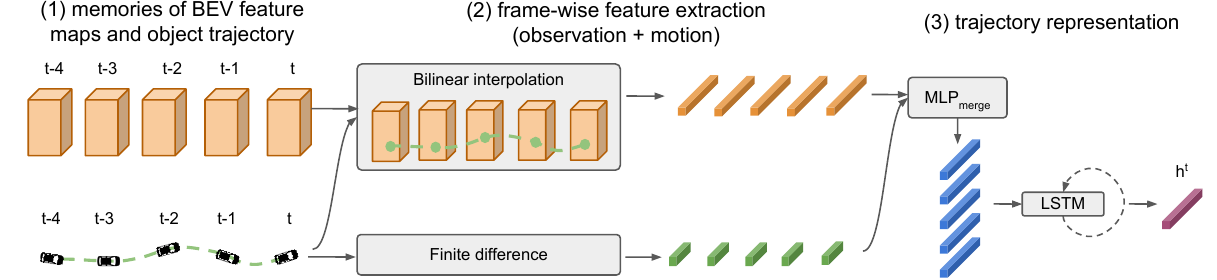}
\end{center}
\caption{\textbf{The proposed trajectory level object representation.} Given an object trajectory, we first extract its sensor observation and motion features at each time step, and then apply an LSTM network to model the temporal dynamics.}
\label{fig:feature}
\end{figure*}

\subsection{Object Detection Module} 
We adopt a 3D object detector that takes multi-sweep LiDAR point clouds (up to 0.5 second) and an HD map as input, and outputs object detections in bird's eye view (BEV). We use a voxel based representation of LiDAR data in BEV, and combine multiple sweeps by concatenating along the height dimension (similar to \cite{intentnet}, with the ego motion compensated for the previous sweeps). We follow \cite{hdnet} to encode the geometric and semantic information of the HD map (if available) into the voxel representation. We apply a 2D convolutional neural network (CNN) based backbone with multi-scale feature fusion to create our intermediate feature representation that will be later used for tracking and motion forecasting
\begin{equation}
\mathcal{F}^t_{\text{bev}}(\mathbf{x}^t) = \text{CNN}_{\text{bev}}(\mathbf{x}^t)
\end{equation}
where $\mathbf{x}^t$ is our input composed of multiple LiDAR sweeps (up to frame $t$) and the HD map.
Following the single stage detector  \cite{pixor} we then use a convolutional detection header to output dense detections, each parameterized as $(u^t_i, v^t_i, w_i, l_i, \theta^t_i)$  representing its position, size and orientation in the ego-centric BEV space at frame $t$. Thus
\begin{equation}
\mathcal{D}^t = \text{CNN}_{\text{det}}(\mathcal{F}^t_{\text{bev}})
\end{equation}
where the number of detections $N_t = |\mathcal{D}^t |$ varies per frame. 
While the detection module generates object detections at each frame independently, the tracking module links them through time, which we review next. 

\subsection{Discrete-Continuous Tracking Module}

There exist two distinct challenges in multi-object tracking: the discrete problem of data association and the continuous problem of trajectory estimation \cite{milan2015multi}. While previous methods mostly focus on the discrete problem, we argue that the continuous problem is as important in our application. From the tracking perspective, it helps to prevent association errors (\ie, identity switches) from accumulating through time. From the prediction perspective, it reduces the variance in motion history caused by the localization error of detections.
Towards this goal, we propose a two-stage tracking framework, where the first stage solves the association problem between previous tracks and current detections, and the second stage refines the associated new tracks to generate smoother trajectories.

\paragraph{Trajectory level object representation:}
We now show how to learn rich and concise representations for the tracking and prediction tasks. 
We formulate the representation learning as a sequence modeling problem (Figure \ref{fig:feature}) and exploit a Long Short-Term Memory (LSTM) network to capture the relevant information. 
Key to the success of the LSTM is to have informed input features. For the task at hand, these features should contain both the object's observation as well as information about its motion. 
Given an object track $\mathcal{P}^{t}_i = \mathcal{D}^{t_0\ldots t}_i$ from frame $t_0$ to frame $t$, 
let $f^{\text{bev},t}_i$ and $f^{\text{velocity},t}_i$ be features representing the observation and motion of each object
\begin{align}
f^{\text{bev, t}}_i & = \text{BilinearInterp}(\mathcal{F}_{\text{bev}}^t, (u_i^t, v_i^t)) \\
f^{\text{velocity}, t}_i & = (\dot{x}_i^t,  \dot{x}_{\text{ego}}^t,  \dot{\theta}_{\text{ego}}^t)
\end{align}
where $\mathcal{F}_{\text{bev}}^t$ is the BEV feature map from the backbone network,  $\dot{x}_i$ and $\dot{x}_{\text{ego}}$ are the 2-dimensional velocities of the $i$-th object and the ego-car respectively, and $\dot{\theta}_{\text{ego}}$ is the angular velocity of the ego-car. 
Note that we estimate the velocities by finite differences over positions, and we use the velocities of each object and the ego car so that we can estimate the absolute velocities.
For newborn objects we initialize its velocity to 0.
For angular velocity of ego car we parameterize it as its cosine and sine values.
We then combine the two features into a single feature representation
\begin{equation}
\label{det_feature}
f(\mathcal{D}_i^t) = \text{MLP}_{\text{merge}}(f^{\text{bev},t}_i, f^{\text{velocity},t}_i)
\end{equation}
The combined object feature is computed for each frame from $t_0$ to $t$, and they are fed to an LSTM network to produce the trajectory level representation
\begin{equation}
\label{tck_feature}
h(\mathcal{P}^{t}_i) = \text{LSTM}(f(\mathcal{D}^{t_0\ldots t}_i))
\end{equation}
We use an LSTM as our sequence model due to its capability to handle varying input length and capture long term dependencies. 
Note that PnPNet exploits the learned trajectory level representation to perform both tracking and prediction tasks. 

\paragraph{Data association:}
Given $N_t$ detections in the current frame and $M_{t-1}$ object tracks in the previous frame, the discrete tracker needs to determine the association between the previous tracks and the current detections. 
In practice we find that the association problem eases when given 3D motion clues. However, properly handling newborn objects and occluded objects can be challenging. Unfortunately both cases happen frequently in driving scenarios.
To handle these two challenges we propose a hybrid approach that exploits the best of multi-object tracking and single-object tracking approaches. 

We first determine the identities of the $N_t$ detections by associating them with all $M_{t-1}$ existing tracks. 
The association problem is formulated as a bipartite matching problem so that exclusive 
track-to-detection correspondence is guaranteed. Newborn objects are handled by adding $N_t$ \textit{virtual candidates} to the $M_{t-1}$ tracks. Note that the result of the association will be fully determined given the affinity matrix that captures the similarity between each detection and track. 
Here, we exploit the learned object representation to compute the affinity matrix $C\in\mathbb{R}^{N_t\times (M_{t-1}+N_t)}$ as follows
\begin{equation}
\label{eq:affinity}
C_{i, j} = 
\begin{cases}
  \text{MLP}_{\text{pair}}(f(\mathcal{D}^t_i), h(\mathcal{P}^{t-1}_j)) & \text{if } 1\le j\le M_{t-1},\\
  \text{MLP}_{\text{unary}}(f(\mathcal{D}^t_i)) & \text{if } j=M_{t-1}+i,\\
  -\text{inf} & \text{otherwise}
\end{cases}
\end{equation}
where $f$ and $h$ are the aforementioned single-frame object feature (Eq. \ref{det_feature}) and trajectory level object feature (Eq. \ref{tck_feature}) respectively.
$\text{MLP}_{\text{pair}}$ computes the affinity score of any detection-track pair, and $\text{MLP}_{\text{unary}}$ estimates the score of any detection being a new instance.
We optimally solve the bipartite matching problem defined by $C$ with the Hungarian algorithm \cite{kuhn1955hungarian}.

Note that dealing with occluded objects via Hungarian matching is very difficult since it is unclear what object estimations should be added to the detection set of the bipartite graph as they are missed by the detector.
To handle such cases we take advantage of single-object tracking (SOT) that is performed on the unmatched tracks (which means the object exists in the past but fails to find a matched observation at current frame).
Our SOT design inherits the philosophy of the Siamese tracker \cite{bertinetto2016fully}, but replaces the correlation filter with a learnable MLP.
Specifically, for each unmatched track $\mathcal{P}^{t-1}_j$, we define its detection candidates $\thicktilde{\mathcal{D}^t_i}$ as voxels within a local neighborhood $\Omega_j$ centered at $(\thicktilde{u}^{t}_j, \thicktilde{v}^{t}_j)$ (estimated by transforming $(u^{t-1}_j, v^{t-1}_j)$ to current frame $t$ with ego motion compensation).
We find the best detection candidate $\thicktilde{\mathcal{D}^t_k}$ by solving for the best match
\begin{equation}
\label{eq:sot}
k = \argmax_{i \in \Omega_j} \text{MLP}_{\text{pair}}(f(\thicktilde{\mathcal{D}^t_i}), h(\mathcal{P}^{t-1}_j))
\end{equation}
In practice, we set the neighborhood size according to the prior knowledge of the object's maximum velocity (110 km/h for vehicles in our case).
Compared with method \cite{weng2019baseline} that predicts the position of occluded object with a motion model, our SOT approach exploits additional observation (such as map context) to get a more precise estimation.

Combining the results from bipartite matching and SOT gives us the final set of tracks $\mathcal{P}^{t}$, which has $N_t+K_t$ instances, where $K_t$ is the number of unmatched tracks that are processed by our single-object tracker. 
Note that all affinity scores in data association are predicted from learnable representations and matching functions, which can learn from data to capture the complex correlations in temporal motion and appearance clues for long-term tracking.

\paragraph{Trajectory estimation:}
The goal of this module is to re-estimate each object track (in terms of the confidence score and trajectory waypoints) given the new observation at current frame, which helps to eliminate false positives from the detector and reduce the localization error coming from either detection or association.
Specifically, for each object track we update its LSTM representation according to the current association, and estimate its confidence score and center position offsets for the most recent $T_0$ frames
\begin{equation}
\label{eq:refine}
\text{score}_i,\Delta u^{t-T_0+1:t}_i,\Delta v^{t-T_0+1:t}_i = \text{MLP}_{\text{refine}}(h(\mathcal{P}^{t}_i))
\end{equation}
$T_0$ is typically shorter than the full trajectory horizon, as near-term history is more relevant to the current frame. After applying the refinement to all tracks, we perform Non-Maximum Suppression (NMS) on current frame estimations ranked by the new scores and keep top $M_t$ tracks to remove false positives and duplicates.

\subsection{Motion Forecasting Module}

While previous joint perception and prediction models \cite{dpt,intentnet} make the prediction module another convolutional header on top of the detection backbone network, which shares the same features with the detection header, in PnPNet we put the prediction module after explicit object tracking, with the object trajectory representation as input
\begin{equation}
\label{eq:pred}
\Delta u^{ t:t+\Delta T}_i,\Delta v^{t:t+\Delta T}_i = \text{MLP}_{\text{predict}}(h(\mathcal{P}^{t}_i))
\end{equation}
where $\Delta T$ is the length of the prediction horizon. 

\subsection{End-to-End Learning}
\input{tables/nusc_det}
\input{tables/nusc_tck}
\input{tables/pnp_eval}

We train our PnPNet end-to-end with a multi-task loss of detection, tracking and prediction:
\begin{equation}
\mathcal{L} = \mathcal{L}_{\text{detect}} + \mathcal{L}_{\text{track}} + \mathcal{L}_{\text{predict}}
\end{equation}
For detection, we use a cross-entropy loss with hard negative mining for classification and sum of smooth $\ell_1$ losses over the bounding box regression terms: size, position and orientation.
For discrete-continuous tracking, we propose to use the max-margin loss on the affinity matrix (Eq. \ref{eq:affinity}), SOT matching scores (Eq. \ref{eq:sot}) and trajectory scores (Eq. \ref{eq:refine}) respectively:
\begin{align}
\mathcal{L}_{\text{track}} & = \mathcal{L}^{\text{affinity}}_{\text{score}} +\mathcal{L}^{\text{sot}}_{\text{score}} + \mathcal{L}^{\text{refine}}_{\text{score}} + \mathcal{L}^{\text{refine}}_{\text{reg}} \\
\mathcal{L}_{\text{score}} & = \frac{1}{N_{i,j}}\sum_{i\in\text{pos},j\in\text{neg}}\text{max}(0, m - (a_i - a_j))
\end{align}
where $a_i$ is the score of $i$-th positive sample, $a_j$ is the score of $j$-th negative sample, $m$ is the margin threshold, and $N_{i,j}$ denotes the number of positive-negative pairs. 
For $\mathcal{L}^{\text{affinity}}_{\text{score}}$ and $\mathcal{L}^{\text{sot}}_{\text{score}}$, we use the pairs of the positive match and each negative matches. For $\mathcal{L}^{\text{refine}}_{\text{score}}$, we use all object pairs (ordered) in which the first object has a larger IoU with the corresponding ground-truth. In this way, the refine score is trained to be ordered by their IoU with the ground-truth. NMS based on this refine score enables higher quality tracks to be kept when there are duplicates.
We set the margin to 0.2 for all scores. 
We use smooth $\ell_1$ loss for both trajectory refinement (Eq. \ref{eq:refine}) and motion forecasting (Eq. \ref{eq:pred}). 

Optimizing PnPNet is nontrivial due to the complex dependencies of the intermediate results across tasks and time. Normal training technique for sequence models like ``teacher forcing" brings exposure bias to the model and leads to severe over-fitting. To address this, we fully emulate the testing phase by sampling a mini-batch of video clips during training.
At each frame, the tracking and prediction modules take as input the \textit{online estimations} from either previous modules or previous frames, and the \textit{ground-truth labels} are only used in computing the multi-task loss. 

We use Adam optimizer \cite{adam} to train PnPNet, with a frame rate of 10 Hz. At each frame we maintain at most $M=50$ tracks and $N=50$ detections per class. The NMS threshold on detections and tracks is 0.1 IoU. We refine the most recent $T_0=4$ frames and predict future $\Delta T=3$ seconds with 0.5 second interval. 
For real-time efficiency, we limit the track length to $T=16$ frames.

%% file: tables/nusc_det.tex

\begin{table}[t]
\begin{center}
\footnotesize
\begin{tabular}{lccccc}
\shline
Methods & \bd{AP $\uparrow$} & AP@0.5m & @1m & @2m & @4m \\
\hline
Mapillary \cite{mapillary} & 47.9 & 10.2 & 36.2 & 64.9 & 80.1 \\
PointPillars \cite{pointpillars} & 70.5 & 55.5 & 71.8 & 76.1 & 78.6 \\
Megvii \cite{zhu2019class} & 82.3 & 72.9 & 82.5 & 85.9 & \bd{87.7} \\
PnPNet, det only & \bd{82.7} & \bd{73.7} & \bd{83.3} & \bd{86.2} & 87.5\\
\shline
\end{tabular}
\end{center}
\caption{\bd{Evaluation of 3D object detection (\textit{car}) on nuScenes.}}
\label{tab:nusc_det}
\end{table}

%% file: tables/nusc_tck.tex

\begin{table*}[t]
\begin{center}
\footnotesize
\setlength\tabcolsep{5pt}
\begin{tabular}{lcccccccccccc}
\shline
Methods & \bd{AMOTA$\uparrow$} & AMOTP$\downarrow$ & RECALL$\uparrow$ & MOTA$\uparrow$ & MOTP$\downarrow$ & MT$\uparrow$ & ML$\downarrow$ & FP$\downarrow$ & IDS$\downarrow$ & FRAG$\downarrow$ & TID$\downarrow$ & LGD$\downarrow$ \\
\hline
StanfordIPRL-TRI \cite{chiu2020probabilistic} & 73.5\% & 0.53 & 73.8\% & 62.3\% & 0.26 & 1978 & 1053 & \bd{6340} & 367 & 341 & 0.79 & 1.08 \\
PnPNet, KF tracker & 76.1\% & 0.52 & 79.1\% & 64.8\% & \bd{0.24} & 2351 & \bd{745} & 7555 & 802 & 628 & 0.51 & 0.97 \\
PnPNet & \bd{81.5\%} & \bd{0.44} & \bd{81.6\%} & \bd{69.7\%} & 0.26 & \bd{2518} & 804 & 6771 & \bd{152} & \bd{310} & \bd{0.30} & \bd{0.57} \\
\shline
\end{tabular}
\end{center}
\caption{\bd{Evaluation of multi-object tracking (\textit{car}) on nuScenes.} Besides standard MOT metrics \cite{bernardin2006multiple}, four new metrics are added: \bd{AMOTA/AMOTP}: MOTA/MOTP averaged over different recall thresholds; \bd{TID}: average track initialization duration in seconds; \bd{LGD}: average longest gap duration in seconds.}
\label{tab:nusc_tck}
\end{table*}

%% file: tables/pnp_eval.tex

\begin{table*}[t]
\begin{center}
\footnotesize
\begin{tabular}{llllllllllll}
\shline
 & \multicolumn{5}{c}{Perception Metrics $\uparrow$} & & \multicolumn{5}{c}{Prediction Metrics $\downarrow$} \\
 & \multicolumn{2}{c}{AP (\%)} & & \multicolumn{2}{c}{Max. Recall (\%)} & & \multicolumn{2}{c}{ADE (m)} & & \multicolumn{2}{c}{FDE (m)} \\
\cline{2-3}\cline{5-6}\cline{8-9}\cline{11-12}
 & \multicolumn{1}{c}{0.1 IoU} & \multicolumn{1}{c}{0.5 IoU} & & \multicolumn{1}{c}{0.1 IoU} & \multicolumn{1}{c}{0.5 IoU} & & \multicolumn{1}{c}{60\% TP} & \multicolumn{1}{c}{90\% TP} & & \multicolumn{1}{c}{60\% TP} & \multicolumn{1}{c}{90\% TP} \\

\rowcolor{grey}\multicolumn{12}{l}{\bt{nuScenes cars}}\\
PnPNet, w/o track & 84.9 & 79.8 & & 90.9 & 84.6 & & 0.69 & 0.75 & & 1.09 & 1.14 \\
PnPNet & 87.1 \sbd{+2.2} & 82.1 \sbd{+2.3} & & 95.3 \sbd{+4.4} & 88.4 \sbd{+3.8} & & 0.58 \sbd{-15\%} & 0.68 \sbd{-9\%} & & 0.93 \sbd{-14\%} & 1.04 \sbd{-8\%} \\

\rowcolor{grey}\multicolumn{12}{l}{\bt{ATG4D vehicles}}\\
PnPNet, w/o track & 93.9 & 90.0 & & 97.5 & 93.4 & & 0.69 & 0.77 & & 1.12 & 1.21 \\
PnPNet & 95.8 \sbd{+2.0} & 92.2 \sbd{+2.2} & & 99.1 \sbd{+1.6} & 95.4 \sbd{+2.1} & & 0.55 \sbd{-20\%} & 0.65 \sbd{-16\%} & & 0.92 \sbd{-18\%} & 1.03 \sbd{-15\%} \\

\rowcolor{grey}\multicolumn{12}{l}{\bt{ATG4D pedestrians}}\\
PnPNet, w/o track & 77.7 & 69.0 & & 88.3 & 78.5 & & 0.39 & 0.41 & & 0.57 & 0.60 \\
PnPNet & 79.5 \sbd{+1.8} & 70.9 \sbd{+1.9} & & 91.0 \sbd{+2.7} & 81.0 \sbd{+2.5} & & 0.34 \sbd{-13\%} & 0.36 \sbd{-11\%} & & 0.51 \sbd{-11\%} & 0.54 \sbd{-10\%} \\
\shline
\end{tabular}
\end{center}
\caption{\bd{Evaluation of end-to-end perception and prediction on nuScenes and ATG4D.} The baseline model (PnPNet, w/o track) follows the paradigm of \cite{dpt}, which performs joint detection and prediction without tracking in the loop.}
\label{tab:pnp_eval}
\end{table*}

%% file: experiments.tex
\section{Experiments}

We demonstrate the effectiveness of PnPNet on two large-scale real-world driving datasets. 
We focus on modular metrics on detection and tracking, as well as system metrics on end-to-end perception and prediction. While modular metrics compare our method with other state-of-the-arts under the constrained setting, system metrics reveal the model performance under the real-world setting.
We show that with the proposed trajectory representation and discrete-continuous tracking, results on each subtask as well as the whole system improve significantly. 
We also provide ablation study of each component and qualitative results of the model.

\subsection{Datasets and Metrics}
\paragraph{nuScenes \cite{nuscenes2019}:}
This dataset contains 1000 20-second log snippets, with 32-beam LiDAR sweeps at 20 Hz and corresponding 3D object labels (linearly interpolated from 2 Hz annotations). 
We train a LiDAR only PnPNet model on nuScenes due to some alignment issues in the maps \footnote{As of map version 1.0. Most issues are resolved in map version 1.2.}. We evaluate on the car class following the official train/val split. 

\paragraph{ATG4D:}
Although nuScenes dataset contains 1000 snippets, they come from 84 unique driving journeys only. The object labels are also constrained to be within 50 meters range, with 63.5\% of the cars being parked. 
In order to better evaluate the real-world performance, especially in urban areas, we evaluate also on the more challenging driving dataset ATG4D \cite{pixor}.
Specifically, ATG4D contains $\sim$5000 log snippets from $\sim$1000 unique journeys in North America. Each snippet has 64-beam LiDAR sweeps at 10 Hz with corresponding HD maps (drivable area, lane graph, and ground height) and 3D object labels within 100 meters range (48.1\% of the cars are parked). We split 500 snippets out for evaluation without journey overlap with the training data.
We train a LiDAR+map PnPNet model and evaluate on the vehicle and pedestrian classes.

\paragraph{Modular metrics:}
We simply follow the detection and tracking metrics defined by nuScenes \cite{nuscenes2019} for a fair comparison with other state-of-the-arts. Specifically, we use Average Precision (AP) for detection and MOT metrics \cite{bernardin2006multiple} for tracking. Metrics are computed under the constrained setting, where we only evaluate on \textit{visible} objects (with at least 1 LiDAR point observation).

\paragraph{System metrics:}
We define system metrics to evaluate the performance of end-to-end perception and prediction, where prediction is conducted on detections instead of ground-truth labels.
Specifically, for perception we use AP and maximum object recall, and for prediction we use Average Displacement Error (ADE) over 3 seconds (with 1 second interval) and Final Displacement Error (FDE) at 3 seconds. Prediction metrics are computed on True Positive (TP) detections at 0.5 IoU.
To mimic real-world setting, we evaluate on \textit{all} objects (including totally occluded ones, which are critical for safety on a self-driving vehicle).

\subsection{Main Results}

\begin{figure}[t]
\begin{center}
   \includegraphics[width=1.0\linewidth]{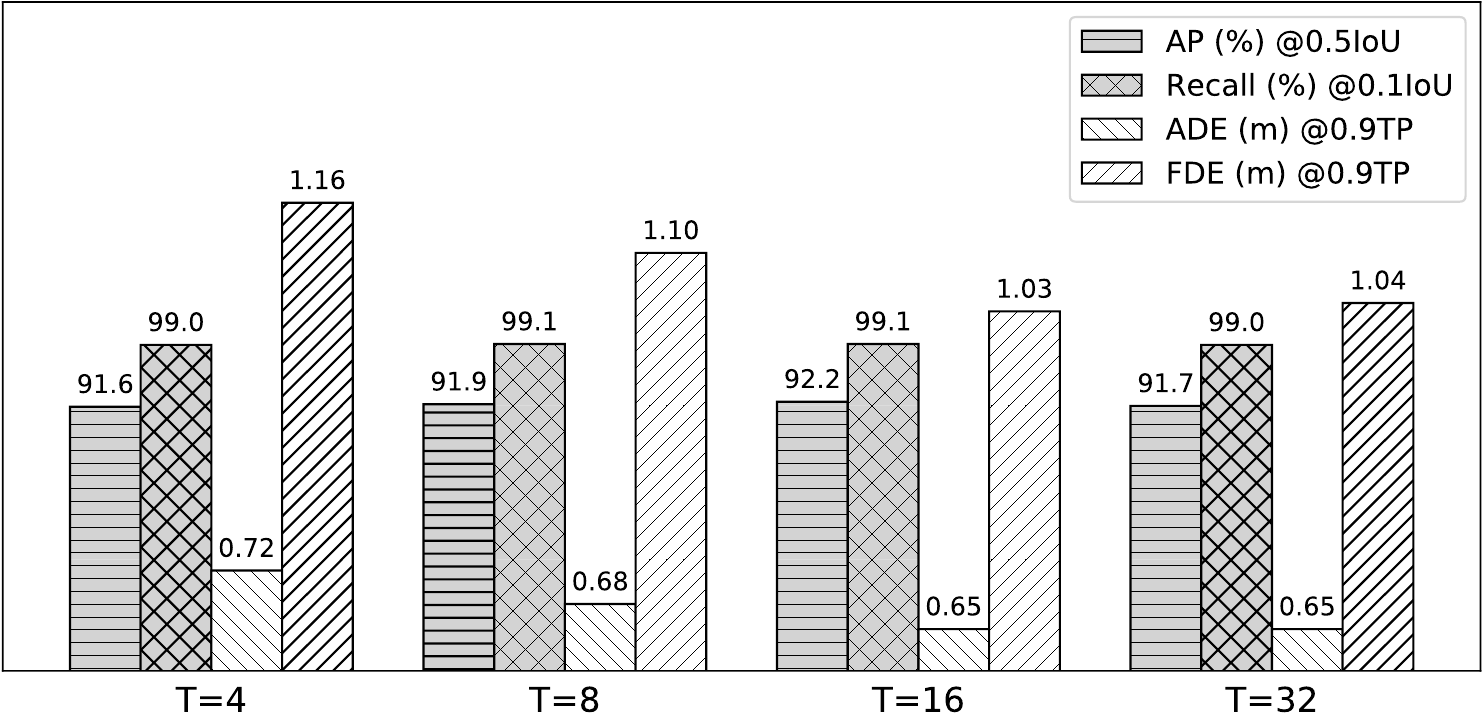}
\end{center}
\caption{\bd{Ablation study on object track length $T$.} Longer track achieves similar perception results but better prediction results. We use T=16 in PnPNet.}
\label{fig:ab-traj-len}
\end{figure}

\paragraph{3D object detection:}
We evaluate the detection module of PnPNet on nuScenes, in comparison with other state-of-the-art 3D detectors. Table \ref{tab:nusc_det} shows that our detector outperforms the leading approach \bd{Megvii} \cite{zhu2019class} in most metrics, with larger gains at higher localization precision (0.8\% AP improvement at 0.5 meter threshold).

\paragraph{Multi-object tracking:}
We evaluate the detection and tracking modules of PnPNet on nuScenes, in comparison with the leading approach \bd{StanfordIPRL-TRI} \cite{chiu2020probabilistic} (with \bd{Megvii} detections \cite{zhu2019class}) on the leaderboard. We also add another tracking baseline that replaces our tracking module with a self-implemented Kalman Filtering based tracker (denoted as ``PnPNet, KF tracker"). Table \ref{tab:nusc_tck} shows that while our KF tracker baseline surpasses \cite{chiu2020probabilistic} by 2.6\% in the ranking metric AMOTA, the proposed PnPNet outperforms \cite{chiu2020probabilistic} by 8.0\%. In terms of fine-grained metrics, PnPNet has more complete trajectories (fewer identity switches and fragmentations), quicker occlusion recovery (smaller track initialization duration and gap duration), and more precise trajectories (smaller AMOTP).

\paragraph{End-to-end perception and prediction:}
Now we evaluate PnPNet in terms of end-to-end perception and prediction on both nuScenes and ATG4D datasets with system metrics under the real-world setting (including totally occluded objects), with an evaluation frame rate of 10 Hz.
We compare with the baseline model that also performs end-to-end perception and prediction, but without tracking in the loop (\ie, we remove the tracking module, and add the prediction header on top of the detection backbone network). We denote this baseline as ``PnPNet, w/o track", which can be considered as a re-implementation of \cite{dpt}. By comparing with this baseline we can measure the effectiveness of the two main contributions of PnPNet, namely the trajectory representation and the discrete-continuous tracking.

Table \ref{tab:pnp_eval} shows that PnPNet consistently outperforms the baseline in all system metrics on two object classes and two datasets, achieving up to 2\% AP gain (note that AP of PnPNet here is evaluated on tracks), up to 4\% recall gain, and up to 20\% prediction improvement. The consistent improvements in different object classes and sensor configurations showcase the generality of the proposed method. More specifically, perception-wise, PnPNet is able to recover from long-term occlusion thanks to the proposed tracking module, which is revealed by up to 4\% boost in recall at 0.1 IoU.
Besides occlusion recovery, PnPNet also benefits from trajectory estimation (for both confidence scores and waypoints), suggested by around 2\% absolute gain in AP at 0.5 IoU.
Prediction-wise, PnPNet achieves 8\% to 20\% relative improvements, which mainly come from two aspects: better perception results from tracking, and stronger object representation at trajectory level. In particular, the gain becomes larger at lower recall (60\% TP) when the perception results are more confident and precise, and still remains significant at 90\% TP where the perception results are noisy. 
We also observe larger gains on ATG4D dataset compared with nuScenes due to larger proportion of moving objects.

\subsection{Ablation Study}

\input{tables/ablation}

\begin{figure*}[t]
\begin{center}
   \includegraphics[trim={0 2.5cm 0 2.5cm},clip,width=0.33\linewidth]{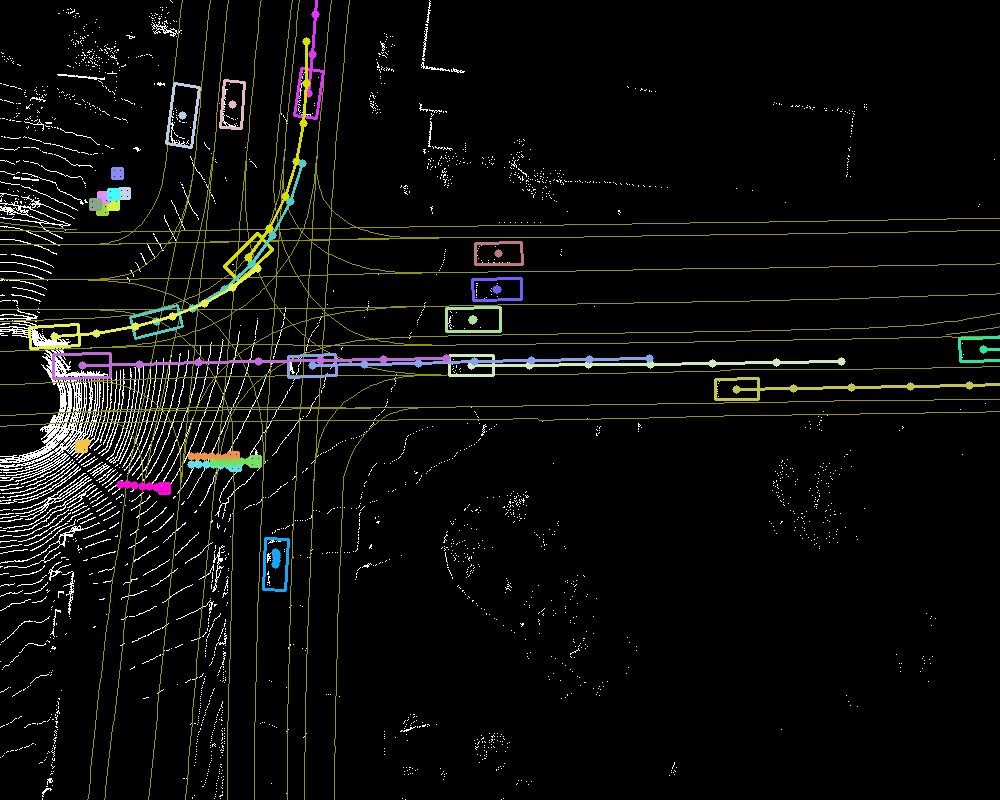}
   \includegraphics[trim={0 2.5cm 0 2.5cm},clip,width=0.33\linewidth]{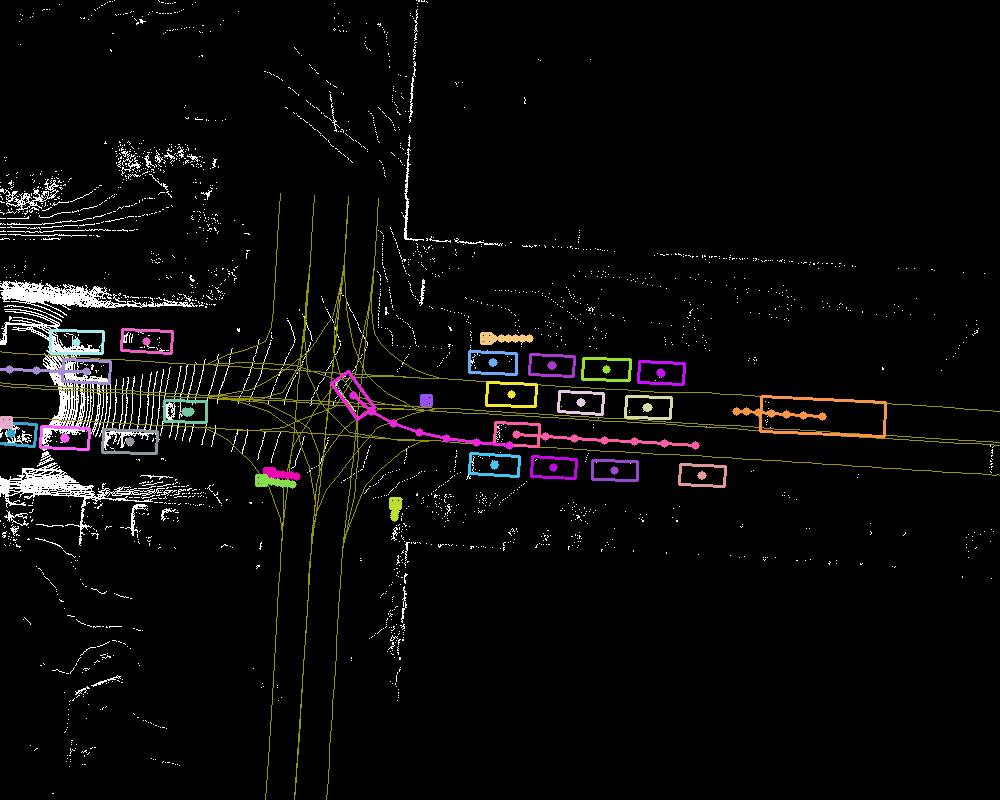}
   \includegraphics[trim={0 2.5cm 0 2.5cm},clip,width=0.33\linewidth]{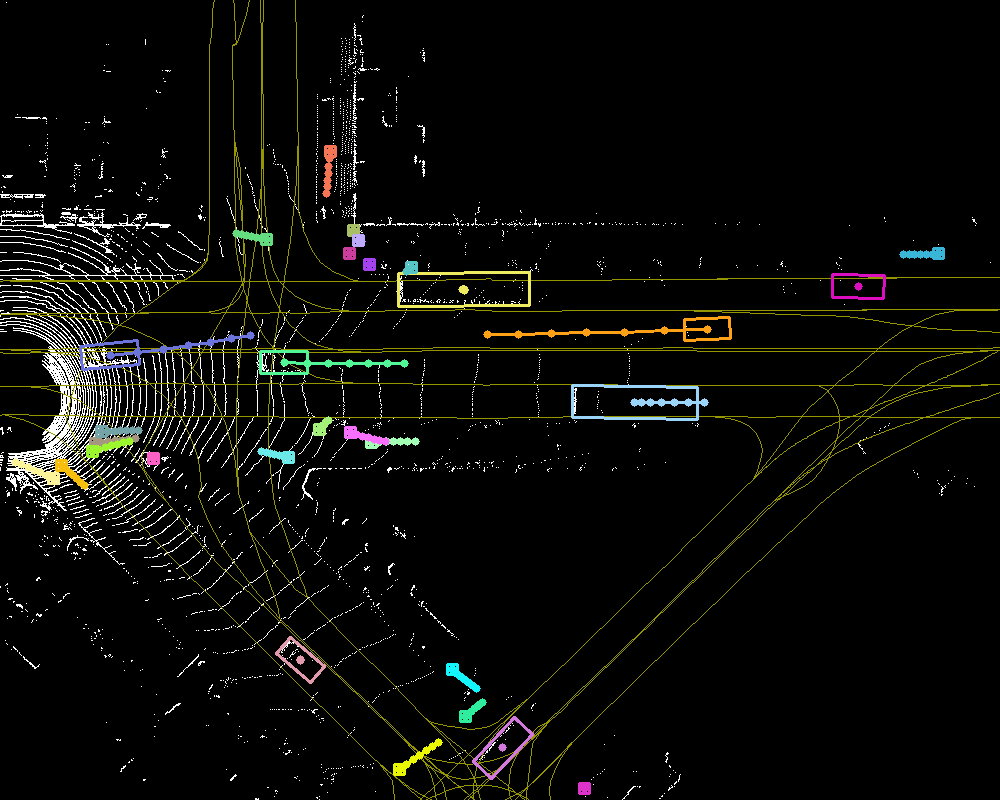}
   \includegraphics[trim={0 2.5cm 0 2.5cm},clip,width=0.33\linewidth]{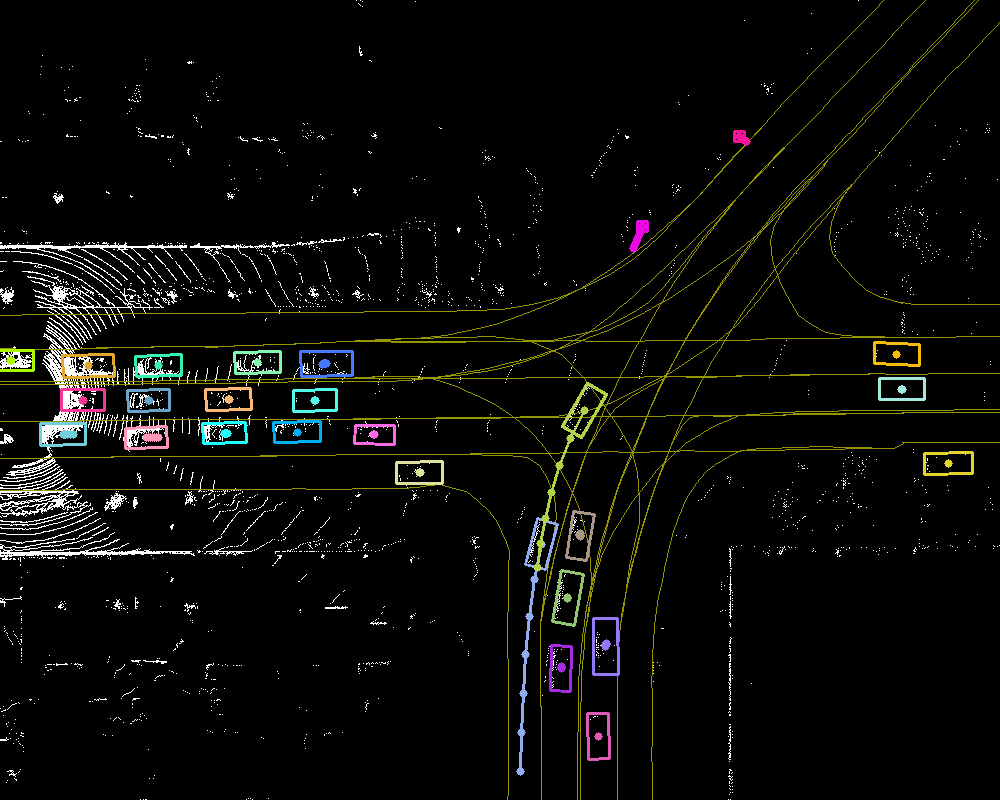}
   \includegraphics[trim={0 2.5cm 0 2.5cm},clip,width=0.33\linewidth]{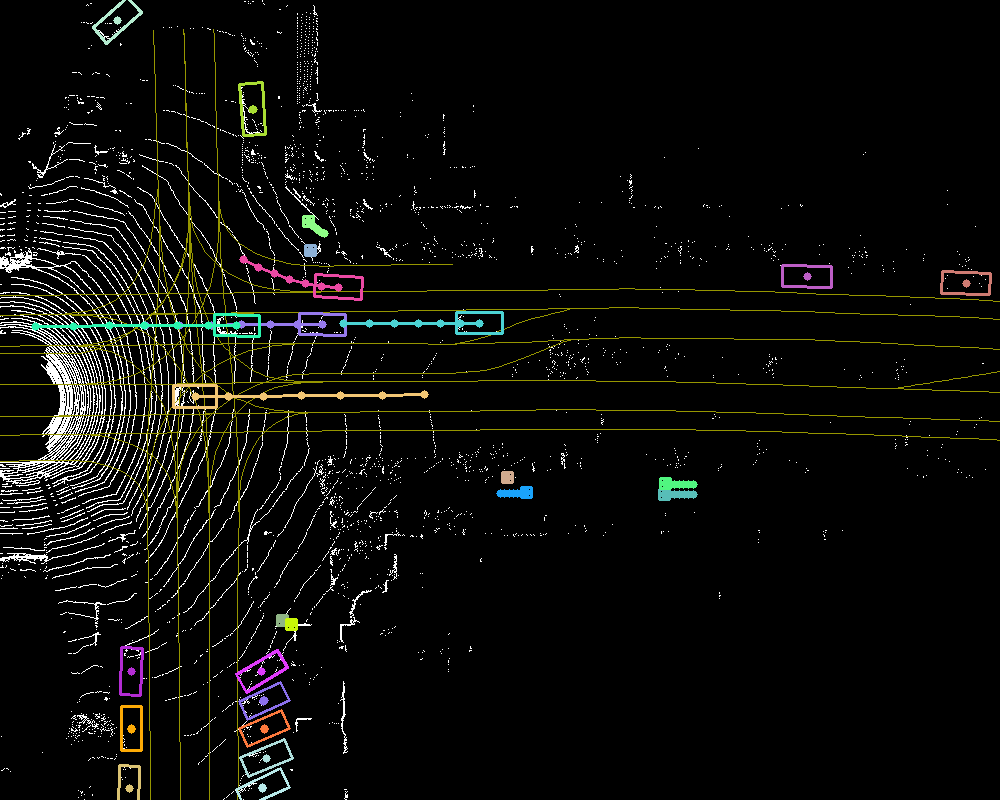}
   \includegraphics[trim={0 2.5cm 0 2.5cm},clip,width=0.33\linewidth]{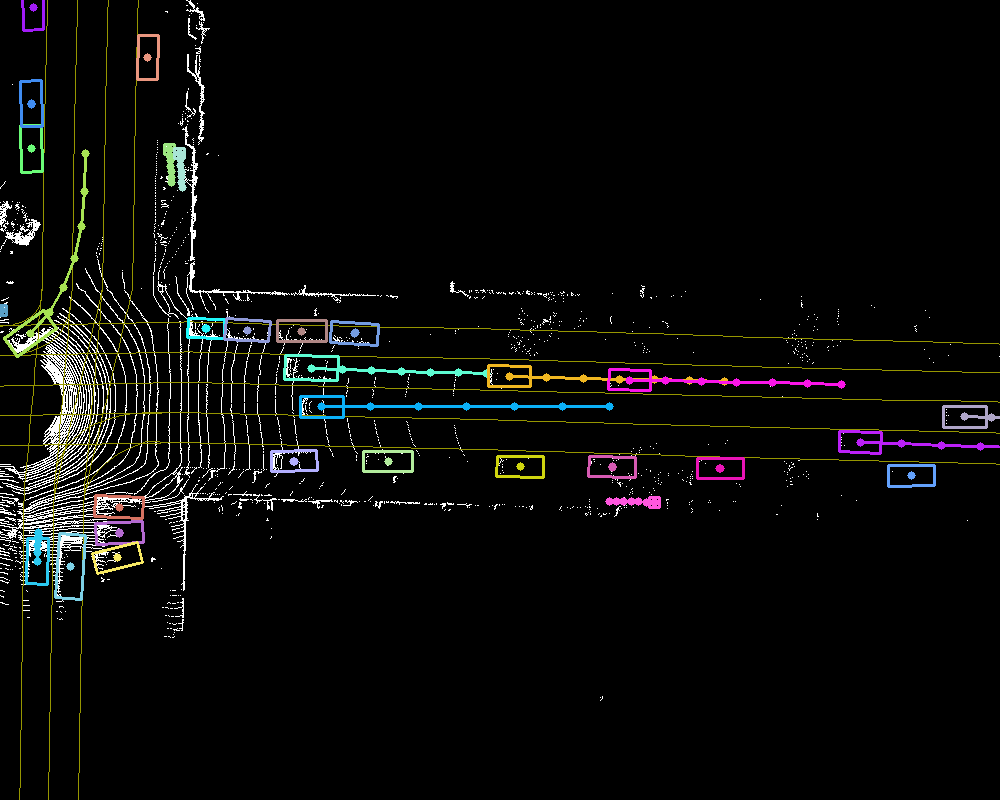}
\end{center}
\caption{\bd{Qualitative results of PnPNet on ATG4D.} We visualize the perception and prediction results of vehicles and pedestrians up to 100 meters far away, where the ego car is located at the middle left of each frame heading to the right.}
\label{fig:demo}
\end{figure*}

We conduct ablation studies on two key components of PnPNet: the object trajectory representation and the discrete-continuous tracker. Note that all ablations are evaluated on ATG4D vehicles with system metrics.

\paragraph{Object track length:} 
We compare PnPNet using different lengths of object track in Figure \ref{fig:ab-traj-len}. From the results we see the history length does not affect perception performance much, suggesting that perception relies more on short-term observations. However, longer track does achieve lower prediction errors, which indicates that long-term history is helpful to future prediction. Prediction performance plateaus at around 16 frames (1.6 seconds), as real-world traffic changes often. 

\paragraph{Importance of explicit motion:} 
One strong finding of PnPNet that was not exploited in  previous joint models \cite{dpt,intentnet} is the fact that exploiting motion from explicit object trajectories is more accurate than inferring motion from the features computed from the raw sensor data. We verify this by removing the motion feature from the trajectory representation of PnPNet, with other components unchanged. As shown in Table \ref{tab:ablation},  the detection performance remains almost the same, but the prediction error increases significantly ($\sim$6\%). This suggests that the explicit motion history obtained from tracking is helpful for prediction.

\paragraph{Single-object tracking for occlusion recovery:}
PnPNet recovers from object occlusion by tracking existing tracks through time. We implement this with a single-object tracker so that current frame's information (\eg, map context) is leveraged as well. If this capability is removed from PnPNet, we observe a performance drop in both perception and prediction (see Table \ref{tab:ablation}). In particular, in the absence of the single-object tracker the recall drops by 1.7\% due to object occlusion, and prediction errors increase by 2\% due to incomplete motion history.

\paragraph{Effect of trajectory estimation:} 
In addition to solving the data association problem in multi-object tracking, PnPNet also re-estimates the trajectory by re-scoring it and refining its waypoints. While re-scoring does not affect the maximum object recall, it determines the order of object trajectories from multiple sources (newborn objects, matched tracks, and tracks through occlusion) and therefore affects the order-dependent metrics. From the results shown in Table \ref{tab:ablation} we can see that, without re-scoring the detection AP drops significantly. Similar performance drop happens in  the prediction metric as well. For trajectory refinement, since it reduces the localization error of online generated perception results, it helps establish a smoother and more accurate motion history. From the results we see that without the trajectory refinement all metrics degrade.

\subsection{Qualitative Results} 

In Figure \ref{fig:demo} we showcase some qualitative results of the proposed model, which illustrate that by learning trajectory representations and explicitly solving multi-object tracking, PnPNet is able to recover from long-term object occlusion, and generate more accurate future trajectories.

%% file: tables/ablation.tex

\begin{table}[t]
\begin{center}
\footnotesize
\setlength\tabcolsep{2pt}
\begin{tabular}{lcccc}
\shline
\multirow{2}{*}{Module removed} & AP (\%) $\uparrow$ & MaxRec. (\%) $\uparrow$ & ADE (m) $\downarrow$ & FDE (m) $\downarrow$\\
 & @0.5IoU & @0.1IoU & 90\%TP & 90\%TP \\
\hline
motion feature & -0.2 & -0.1 & +6.2\% & +5.5\% \\
single object track & -1.7 & -1.7 & +2.0\% & +2.0\% \\
trajectory rescore & -7.9 & -0.4 & +4.7\% & +4.8\% \\
trajectory refine & -2.1 & -1.6 & +4.8\% & +4.7\% \\
\rowcolor{grey}whole track module & -2.2 & -1.6 & +18\% & +17\% \\

\shline
\end{tabular}
\end{center}
\caption{\bd{Ablation study on discrete-continuous tracking.} We remove one module from the full PnPNet with other modules unchanged and report the relative performance change.}
\label{tab:ablation}
\end{table}

%% file: conclusion.tex
\section{Conclusion}

In this paper we proposed PnPNet, an end-to-end model for perception and prediction in autonomous driving.
Instead of designing individual models for each subtask like the traditional engineering stack, we follow the recent advances of joint modeling with shared feature computation, and further improve upon the paradigm with a novel multi-object tracker that generates object trajectories online from detections and exploits trajectory level features for motion forecasting.
We validate PnPNet on two large-scale driving datasets, and show significant improvements in both perception and prediction metrics.
In the future we plan to apply our approach to more complex downstream tasks like multi-agent behavior prediction and motion planning.

%% file: appendix.tex
\clearpage

\begin{figure*}[t]
\begin{center}
   \includegraphics[width=1.0\linewidth]{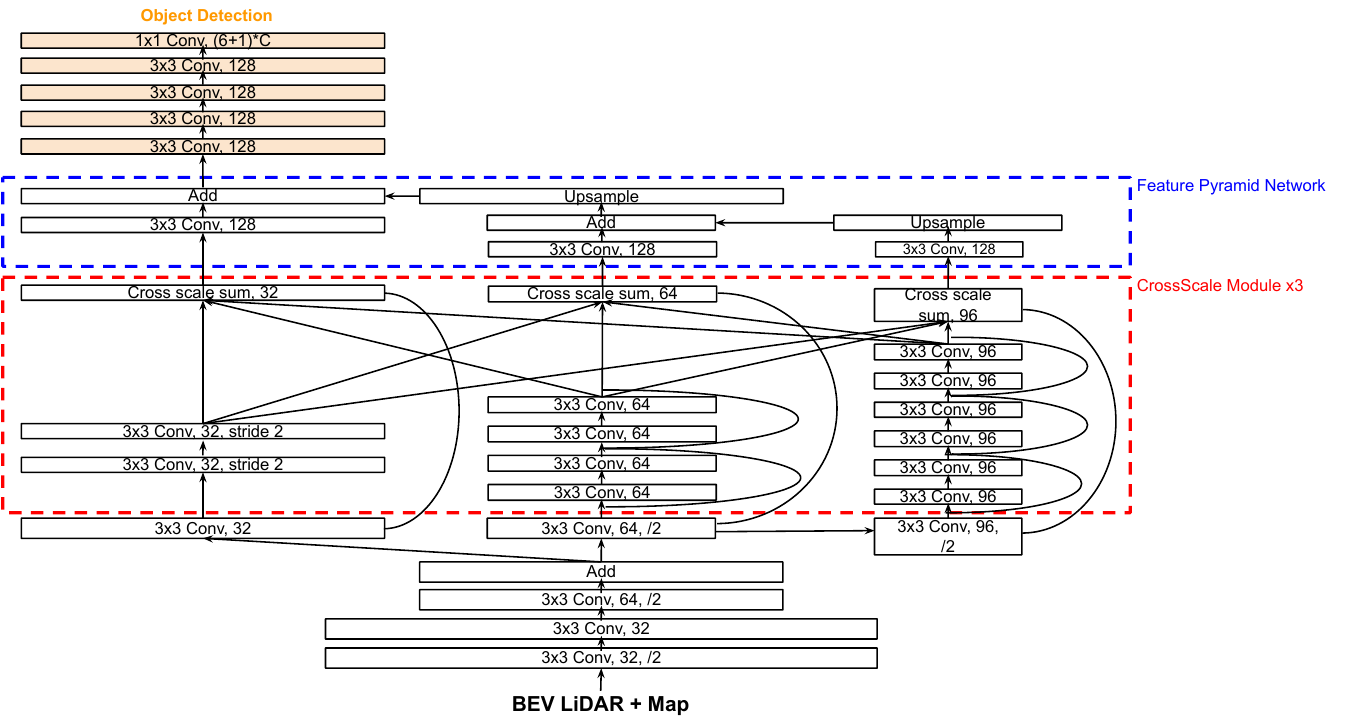}
\end{center}
   \caption{\bd{Architecture of the backbone network.}}
\label{fig:backbone}
\end{figure*}

\begin{figure*}[t]
\centering
\begin{tabular}{ccc}
\includegraphics[width=0.33\linewidth]{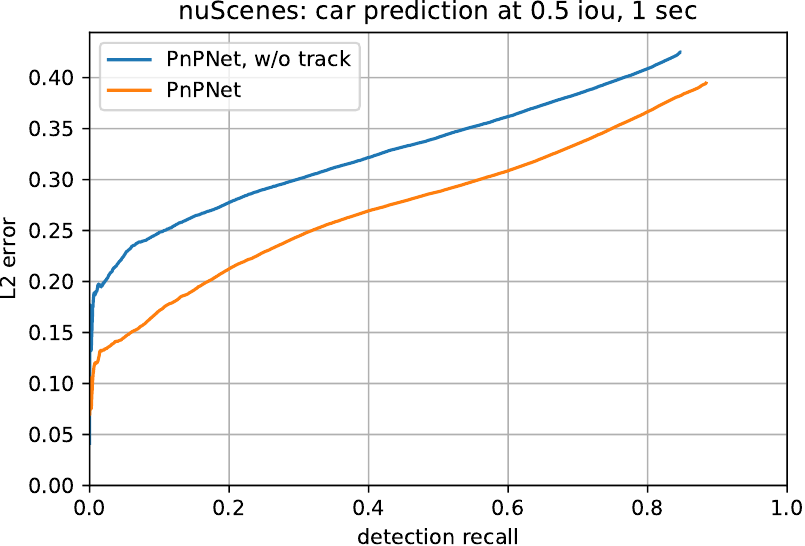} &
\includegraphics[width=0.33\linewidth]{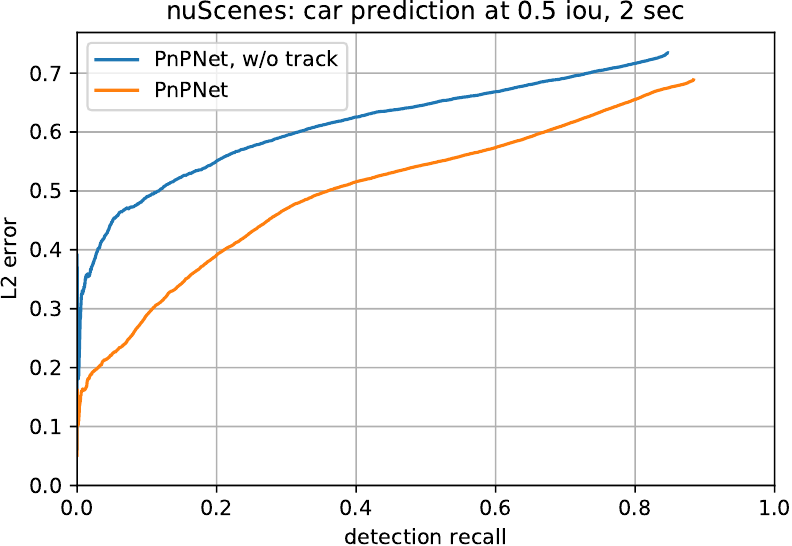} &
\includegraphics[width=0.33\linewidth]{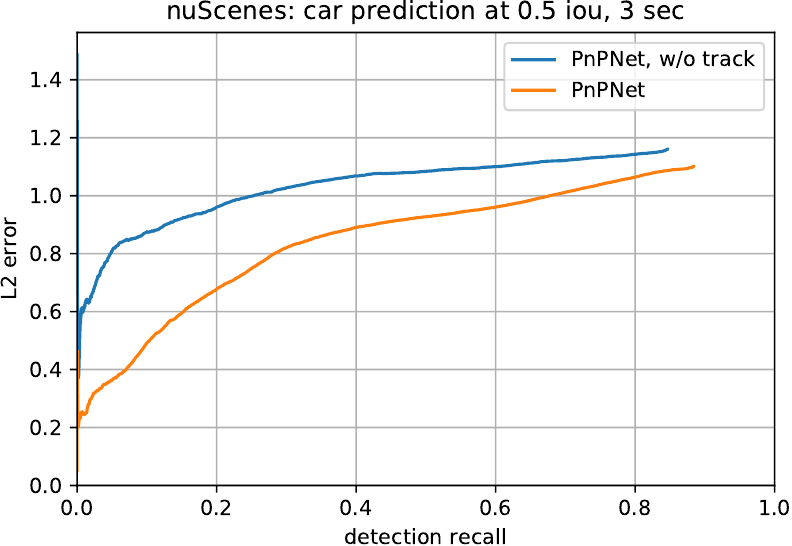} \\
\\
\includegraphics[width=0.33\linewidth]{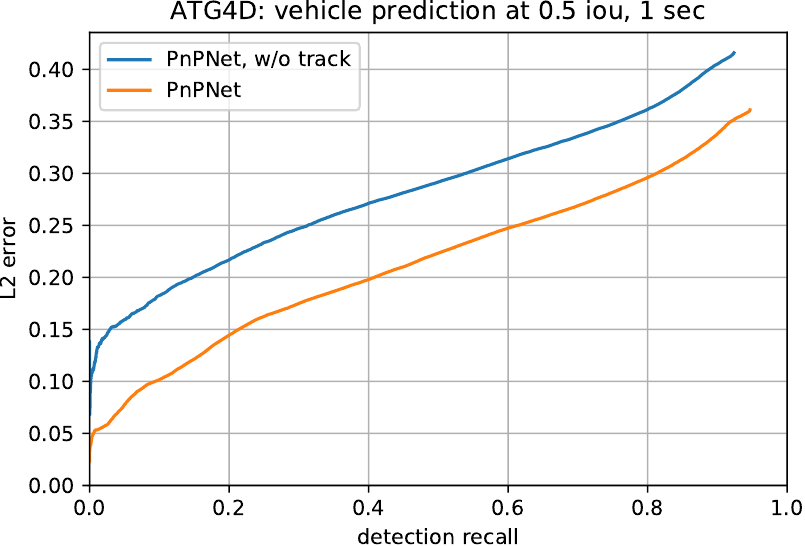} &
\includegraphics[width=0.33\linewidth]{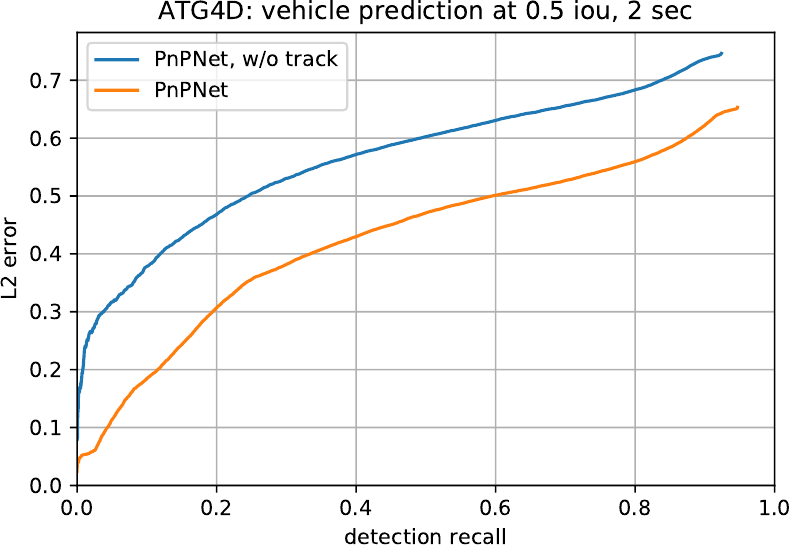} &
\includegraphics[width=0.33\linewidth]{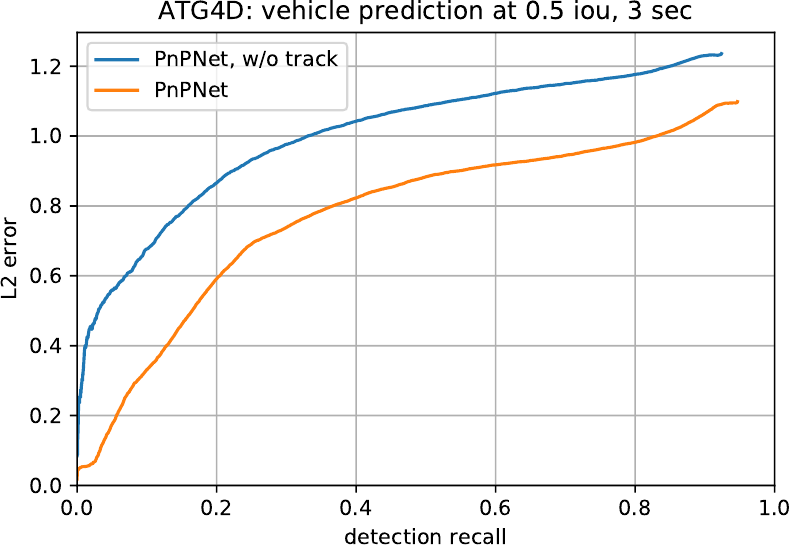} \\
\\
\includegraphics[width=0.33\linewidth]{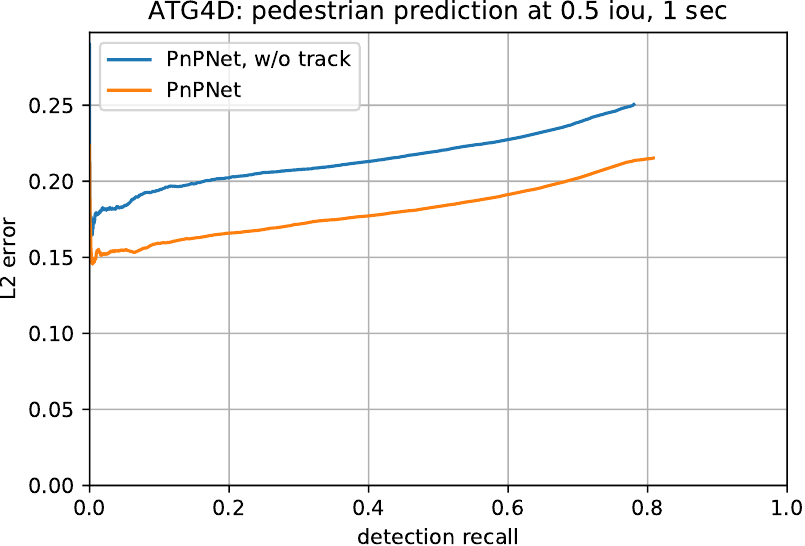} &
\includegraphics[width=0.33\linewidth]{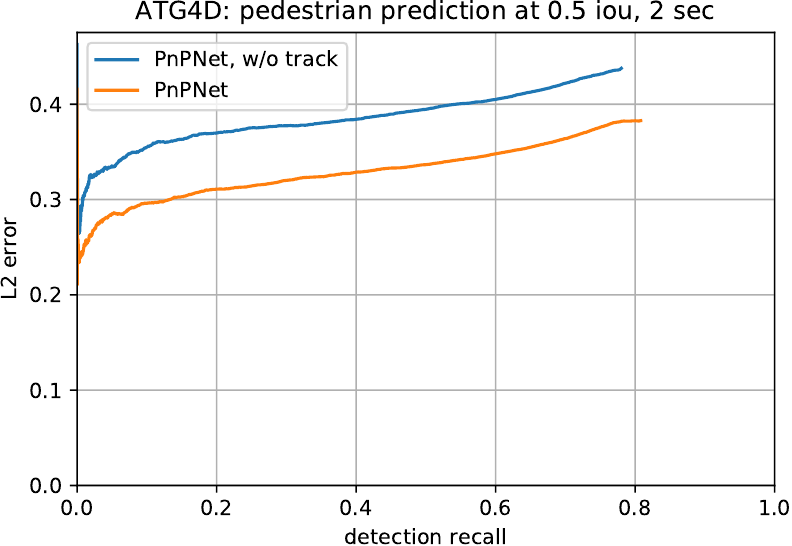} &
\includegraphics[width=0.33\linewidth]{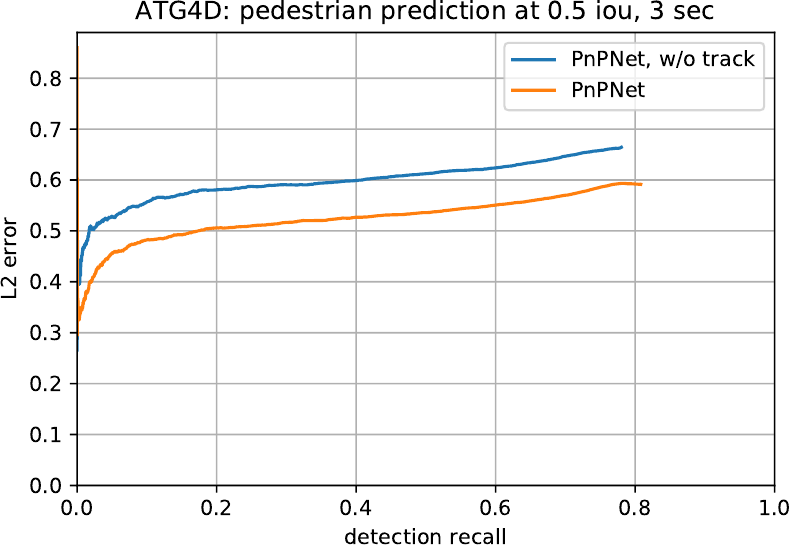} \\
\\
\end{tabular}
\caption{\bd{Final Displacement Error (FDE) at 1s, 2s, 3s prediction for all recall rates on nuScenes and ATG4D datasets.}}
\label{fig:pred_eval}
\end{figure*}

\section*{Appendix}
\section*{A. Backbone Network Architecture}

We first show in Figure \ref{fig:backbone} the architecture of the backbone network. We use bird's eye view (BEV) occupancy map as input representation for the LiDAR, and concatenate the BEV representations of multi-sweep LiDAR point clouds (1 current frame + $N$ previous frames) along the height dimension.  We employ $N=9$ in nuScenes dataset and $N=4$ in ATG4D dataset. For map representation, we exploit both geometric and semantic priors similar to HDNet \cite{hdnet}. Specifically, we do ground height subtraction on the multi-sweep LiDAR point clouds, and compute two additional binary raster images representing the drivable region and the lane graph respectively (for the current frame only). The map raster images are concatenated with the LiDAR BEV representation along the height dimension.

Given the aforementioned BEV representation of both LiDAR and HD maps as input, we first apply three \texttt{Conv2D} layers to down-sample the input BEV images by a factor of $4$. We then apply a cross-scale module sequentially three times. This cross-scale module is inspired by the Inception block with residual connections \cite{szegedy2017inception}. The difference is that feature maps are spanned at multiple scales ($3$ in our case), and each scale receives information from all other scales. This leads to a better trade-off between accuracy and speed. After three cross-scale modules, we apply a feature pyramid network \cite{fpn} to combine multi-scale feature maps, resulting in a 4$\times$ down-sampled BEV feature map with $128$ channels. For the detection header we simply use $4$ \texttt{Conv2D} layers each with $128$ filters. The detection header outputs $(6+1)*C$ channels as dense detection estimations, which correspond to $(x,y,w,l,\sin\theta,\cos\theta,\text{score}\_\text{logit})$ for each object category.

\section*{B. Implementation Details on nuScenes}

We use the same network architecture on both nuScenes and ATG4D datasets. On nuScenes, following the dataset rule imposed by the creators of the dataset, we aggregate 10 sweeps of LiDAR point clouds (1 current and 9 previous) corresponding to 0.5 seconds of past history. We consider the point clouds within a region of $[-50, 50]\times[-50, 50]\times[-3, 5]$ meters around the ego car. We use a voxel size of $0.15625\times0.15625\times0.25$ meters, leading to a voxel grid size of $640\times640\times320$ as input.

We apply frame-level data augmentation during training. Specifically, labels at non-key frames are linearly interpolated from labels at adjacent key frames. For each frame, we apply random scaling (0.95 $\sim$ 1.05 for all 3 axes), translation (-1 $\sim$ 1 meters for XY axes and -0.2 $\sim$ 0.2 meter for Z axis), rotation (-45 $\sim$ 45 degrees along Z axis) and flipping (along X axis) to both 3D LIDAR point clouds and 3D object labels.

The model is trained on the car class, and we ignore labels that have 0 LiDAR point inside the box or outside the 50 meters range with respect to the ego car. During training we define positive samples as voxels with IoU (assuming ground-truth size and orientation) larger than 0.9, and define negative samples as smaller than 0.4 IoU. We use Adam optimizer and train with batch size of 8 for 4 epochs. The initial learning rate is 0.001, and is decayed by 0.1 at 2.8 and 3.6 epochs respectively.

\section*{C. Fine-Grained Evaluation of Motion Forecasting}

While in the main paper we evaluate prediction errors (ADE and FDE) of motion forecasting at fixed object recall rates, here we provide more fine-grained evaluation of FDE (at 1s, 2s and 3s respectively) for all recall rates. We show evaluation results in Figure \ref{fig:pred_eval}, where we observe that the proposed PnPNet not only achieves higher object detection recall, but also outperforms the baseline in prediction consistently at all recall rates.